\documentclass[journal,twocolumn,transmag]{IEEEtran}
\usepackage{caption}
\usepackage{times}
\usepackage{epsfig}
\usepackage{graphicx}
\usepackage{amsmath}
\usepackage{amssymb}
\usepackage{multirow}
\usepackage{subfigure}
\usepackage{ulem}
\usepackage{color} 
\usepackage{enumerate}

\usepackage[backref]{hyperref}
\usepackage{makecell}

\ifCLASSINFOpdf
\else
\fi
\begin{document}
%
\title{Multiple Information Prompt Learning for Cloth-Changing Person Re-Identification}

\author{Shengxun Wei, Zan Gao, \IEEEmembership{Senior Member, IEEE}, Chunjie Ma, \\ Yibo Zhao, Weili Guan, Shengyong Chen, \IEEEmembership{Senior Member, IEEE, IET Fellow} }


\markboth{}%
{Shell \MakeLowercase{\textit{et al.}}: Bare Demo of IEEEtran.cls for IEEE Transactions on Magnetics Journals}
%




\maketitle
\begin{abstract}
Abstract - Cloth-changing person re-identification is a subject closer to the real world, which focuses on solving the problem of person re-identification after pedestrians change clothes. The primary challenge in this field is to overcome the complex interplay between intra-class and inter-class variations and to identify features that remain unaffected by changes in appearance. Sufficient data collection for model training would significantly aid in addressing this problem. However, it is challenging to gather diverse datasets in practice. Current methods focus on implicitly learning identity information from the original image or introducing additional auxiliary models, which are largely limited by the quality of the image and the performance of the additional model. To address these issues, inspired by prompt learning, we propose a novel multiple information prompt learning (MIPL) scheme for cloth-changing person ReID, which learns identity robust features through the common prompt guidance of multiple messages. Specifically, the clothing information stripping (CIS) module is designed to decouple the clothing information from the original RGB image features to counteract the influence of clothing appearance. The bio-guided attention (BGA) module is proposed to increase the learning intensity of the model for key information. A dual-length hybrid patch (DHP) module is employed to make the features have diverse coverage to minimize the impact of feature bias. Extensive experiments demonstrate that the proposed method outperforms all state-of-the-art methods on the LTCC, Celeb-reID, Celeb-reID-light, and CSCC datasets, achieving rank-1 scores of 74.8\%, 73.3\%, 66.0\%, and 88.1\%, respectively. When compared to AIM (CVPR23), ACID (TIP23), and SCNet (MM23), MIPL achieves rank-1 improvements of 11.3\%, 13.8\%, and 7.9\%, respectively, on the PRCC dataset. \footnote{Manuscript received June-27th, 2024;  This work was supported in part by the National Natural Science Foundation of China (No.62372325, No.62402255), Natural Science Foundation of Tianjin Municipality (No.23JCZDJC00280), Shandong Provincial Natural Science Foundation (No.ZR2024QF020), Shandong Province National Talents Supporting Program (No.2023GJJLJRC-070), Shandong project towards the integration of education and industry (No.801822020100000024), Young Talent of Lifting engineering for Science and Technology in Shandong (No. SDAST2024QTB001), Shandong Project towards the Integration of Education and Industry (No.2024ZDZX11).

S.X Wei, Z. Gao, Y.B Zhao and S.Y Chen are with Key Laboratory of Computer Vision and System, Ministry of Education, Tianjin University of Technology, Tianjin, 300384, P.R China.

C.J Ma and Z. Gao are with the Shandong Artificial Intelligence Institute, Qilu University of Technology (Shandong Academy of Sciences), Jinan, 250014, P.R China.

W.L Guan is with the School of Electronics and Information Engineering, Harbin Institute of Technology, Shenzhen, 150001, P.R China. 
}




\end{abstract}


\maketitle

\IEEEdisplaynontitleabstractindextext

%
\IEEEpeerreviewmaketitle

\section{Introduction}
%
%
%
%

\IEEEPARstart
Person Re-identification (ReID) is a prominent research focus within the fields of computer vision and machine learning. It aims to solve the problem of retrieving target pedestrians across non-overlapping cameras, and determine whether the images taken by different cameras contain a specific pedestrian. This task plays an important role in areas such as public safety, smart commerce, and smart security (e.g., identifying customers for personalization, tracking potential criminal suspects). Traditional person ReID methods \cite{sun2018PCB,wang2018MGN,nguyen2018kernel,ren2019uniform,Gao2021DCR,he2021transreid,Zhou22OS-Net,Ye2022AGW,li2023clip-reid} usually rely on pedestrians wearing the same clothes in different shots. However, in real-world scenarios, pedestrian appearance can change significantly over time, making it difficult for models to learn effective identification features. This variability presents a substantial challenge for person ReID. To address this issue, researchers are increasingly focusing on the cloth-changing person re-identification (CC-ReID) task. This approach aims to identify other identity-related features that are independent of clothing, such as body posture and gait. These features can still effectively identify pedestrians even when their clothing changes.

\begin{figure}[t]
  \centering
  \includegraphics[width=\linewidth]{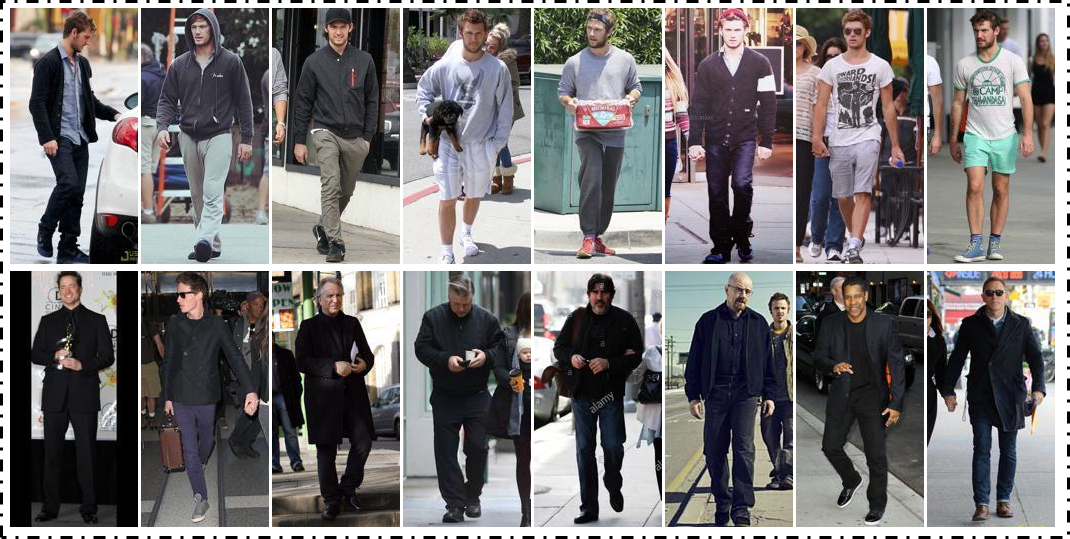}
  \caption{Examples of cloth-changing person ReID images. Intra-class variation and inter-class variation of pedestrian samples in the cloth-changing scenario.}
  \label{fig:example-ccreid}
\end{figure}

Some researchers \cite{Chen20213DSL,Gu2022CAL,Gao2022MVSE,bansal2022VIBE,Yang2023AIM,Guo2023SCNet,Wang2023CSCL} have made active attempts in person ReID task under changing clothes scenario. The public datasets PRCC \cite{yang2021PRCC}, LTCC \cite{qian2020LTCC}, Celeb-reID \cite{huang2020Celeb-reID}, Celeb-reid-light \cite{huang2019Celeb-reID-light} and CSCC \cite{yan2022UCAD} for person ReID research have also been constructed. Figure \ref{fig:example-ccreid} shows some examples of CC-ReID images, where the first row represents the same person wearing different clothes and the second row represents different people wearing similar clothes, from which we can observe that the first row exhibits a large difference in pedestrian appearance and the second row shows a very small difference in pedestrian appearance. Under CC-ReID, the appearance of human clothing can no longer be used as the main basis for recognition but has become an interference factor. At this time, the traditional ReID method that largely depends on the appearance characteristics of clothing is no longer competent, so the ReID method for changing clothes is a more challenging and urgent problem to be solved. In CC-ReID, because of the unreliability of clothing information, the intuitive response is to model identity information unrelated to clothing. Recently, some novel methods for changing person ReID have been proposed, for example, Jin et al. \cite{Jin2022GI-reid} proposed a Gait-assisted Image-based cloth-changing ReID (GI-ReID) framework, which introduces Gait information to assist in learning clothing independent representations. Gao et al. \cite{Gao2022MVSE} proposed a novel multigranular visual-semantic embedding algorithm (MVSE) to deeply explore the visual semantic information and pedestrian attribute information, so as to effectively solve the problem of CC-ReID.

There are two main challenges in the CC-ReID: I) intra-and inter-class variation of people caused by changes in clothing appearance. II) changes in pedestrian pose/viewpoint and pedestrian occlusion. Although previous studies have achieved performance gains in addressing these issues, there are still some limitations: 1) \textbf{Spatial redundancy exists in visual modes}. Existing methods are implemented from the perspectives of adversarial training \cite{Gu2022CAL,Yang2023ACID}, domain generalization \cite{Zhao2023imsgep}, causal intervention \cite{Yang2023AIM}, etc., to learn effective feature representations only from RGB images. However, there is a lot of spatial redundancy in the pure RGB image \cite{he2022mae}, and the extracted independent representation of identity will be greatly affected by the changes of clothing appearance and background. It is difficult to effectively decouple interference features such as clothing, resulting in poor performance of existing methods. Therefore, it is an urgent problem to control the influence of spatial redundant information on its visual representation to improve the discriminative stability of the model. 2) \textbf{Identity clues are underutilized}. Many existing ReID methods \cite{Han2023CCFA} lack the use of key information of pedestrians, and often implicitly learn identity classification features from the original features of pedestrians, and do not make full use of key information related to identity to prompt model learning. Therefore, explicitly prompting the model to learn the key features of identity is worth investigating. 3) \textbf{Feature bias affects performance}. In CC-ReID, features are still biased by pose changes, occlusion and other conditions. The existing methods \cite{Chen20213DSL, Chen2022MAC-DIM} mainly focus on using auxiliary information to overcome the influence of bias, but their effectiveness depends on the quality of auxiliary information and will incur additional resource overhead. Therefore, how to effectively reduce the impact of feature bias is worth exploring.

To address the above issues, we design a new MIPL framework to learn identity robust features using the prompt guidance of multiple information. \textbf{For problem (1)}, inspired by \cite{li2023clip-reid}, we introduce a multi-stage vision-language learning strategy to effectively establish the correspondence between visual representation and high-level language description, and design a clothing information stripping module to effectively decouple clothing information from original image features with the prompt of clothing attribute text description. \textbf{For problem (2)}, the biological information guidance branch is proposed, which explicitly prompts the model to learn biological key features with strong identity correlation through local unique biological information attention prompts. In addition, \textbf{for problem (3)}, a dual-length hybrid patch module is designed to reduce the impact of feature deviation and make the features have diverse coverage. The main contributions of this paper are summarized as follows.
\begin{itemize}
\item We propose a novel MIPL algorithm for cloth-changing person ReID, which introduces a vision-language learning strategy. The prompt-guidance process is optimized in an end-to-end unified framework through the common prompt guidance of multiple information, and makes full use of various information to learn identity robust features.
\item We design a novel clothing information stripping module (CIS), which effectively decouples the clothing information from the original image features and counteracts the influence of clothing appearance. We propose the bio-guided Attention (BGA) module that prompts the model to learn biologically key features whose identities are strongly correlated. We develop a dual-length hybrid patch (DHP) module to minimize the impact of feature biases caused by occlusion, pose variation, and viewpoint differences.
\item We conducted a systematic and comprehensive evaluation of the MIPL algorithm on five public cloth-changing person ReID datasets, and the experimental results show that the MIPL method outperforms the existing cloth-changing person ReID methods in terms of mAP and rank-1.
\end{itemize}

The remainder of the paper is organized as follows. Section II introduces the related work, and Section III describes the proposed MIPL method. Section IV describes the experimental settings and the analysis of the results. Section V presents the details of the ablation study, and concluding remarks are presented in Section VI.

\section{Related Work}

As an important technology of intelligent video surveillance, ReID has attracted the attention of researchers, and many methods have been proposed. According to visual appearance, these methods can be roughly divided into clothing-consistent Person ReID and cloth-changing Person ReID. In the following, we will separately introduce them.

\subsection{Cloth-consistent Person ReID}

In the field of person ReID, the characteristics of commonly used datasets and the limitations of related methods are worthy of in-depth discussion. The datasets Market-1501 \cite{zheng2015Market1501}, MSMT17 \cite{wei2018MSMT17}, and CUHK03 \cite{li2014cuhk} used by most person ReID methods are collected in a short period of time, and they are all carried out under the assumption that the appearance of pedestrians' clothes will not change. These datasets mainly focus on traditional factors such as illumination, pose, viewpoint, and occlusion and achieve satisfactory performance. For instance, relying on the powerful modeling ability of the CNN network \cite{he2016ResNet}, Zhou et al. \cite{Zhou22OS-Net} designed a new omni-scale network (OSNet). This network emphasizes multi-scale feature capture and combination to achieve better representation. However, OSNet may have higher computational complexity due to its multi-scale feature capture design, and there is a risk of overfitting with its intricate architecture. Meanwhile, Ye et al. \cite{Ye2022AGW} offers a comprehensive analysis and new baseline method in person reID. Reviews and analyzes existing person ReID techniques from deep feature representation learning, deep metric learning, and ranking optimization. Gao et al. \cite{Gao2021DCR} proposed a deep spatial pyramid feature collaborative reconstruction model (DCR), which uses joint reconstruction to address the issues of pedestrian pose, perspective changes, and occlusion in ReID. It provides effective support for person ReID tasks in scenarios with these challenges. With the success of Transformers \cite{dosovitskiy2021ViT} in vision tasks, He et al. \cite{he2021transreid} proposed a pure transformer-based baseline framework (TransReID). This framework encodes edge information such as viewpoint and camera through learnable embeddings and rearranges patch embeddings to generate more discriminative features.

These methods can effectively deal with traditional factors such as illumination, pose, perspective and occlusion, but are limited to completing the person ReID task in a short period of time (that is, under the assumption that the person's clothing appearance does not change).

\subsection{Cloth-changing Person ReID}

In the real scenario, the assumption that the appearance of clothes does not change over time is quickly broken, and the cloth-consistent person ReID method is no longer efficient. Therefore, researchers begin to establish some cloth-changing person ReID datasets \cite{huang2019Celeb-reID-light,yu2020cocas,qian2020LTCC,huang2020Celeb-reID,yang2021PRCC,yang2022sampling}, and pay attention to the task of cloth-changing person ReID. For example, Yu et al. \cite{yu2020cocas} constructed a novel change of clothing recognition benchmark COCAS, which uses clothing templates and pedestrian images for combination to search the target image. 
Additionally, in order to reduce the reliance on extensive data collection, Jia et al. \cite{Jia2022Complementary} enhanced the feature learning process by devising a potent complementary data augmentation strategy named Pos-Neg. This strategy jointly prompts the model to learn more robust and discriminative representations without requiring additional information or incurring the cost of expanding the latent sample space. Han et al. \cite{Han2023CCFA} aim to enlarge the training data and put forward a novel clothing-change Feature Augmentation (CCFA) model that implicitly integrates semantically meaningful Clothing Change augmentation in the feature space. Liu et al. \cite{Liu23Dual} proposed a Dual-Level Adaptive Weighting (DLAW) solution to measure the degree of cloth-changing and assess the influence of image and feature levels on cloth-changing patterns, thereby resolving the cloth-changing ReID problem.
Merely relying on data augmentation cannot comprehensively and effectively express the targets in images. Therefore, researchers have become keen on using auxiliary information to enable the model to learn jointly. So, Chen et al. \cite{Chen20213DSL} combined person ReID and 3D human reconstruction to propose an end-to-end architecture for 3D shape learning (3DSL) to extract texture-insensitive 3D shape embeddings directly from 2D images. Subsequently, Hone et al. \cite{Hong2021FSAM} proposed a two-stream Fine-grained shape-appearance Mutual learning (FSAM) framework to supplement clothing-independent knowledge from shape streams to appearance features through dense interactive mutual learning. Moreover, Chen et al. \cite{Chen2022MAC-DIM} proposed a multi-scale appearance and contour deep infomax(MAC-DIM) module to maximize the mutual information between color appearance features and wheel shape features to overcome the problem of clothing color bias. Liu et al. \cite{Liu2023PGAL} proposed a Pose-Guided Attention Learning (PGAL) framework, which uses a pose estimation network to align keypoint features and improves the feature representation of non-keypoint regions of the human body through multi-head self-attention. Zhang et al. \cite{Zhang2023Multi} proposed the Multi-Biometric Unified Network (MBUNet) framework, which leverages multi-biometric features to learn cloth-changing cues that are unrelated to clothing.  Wang et al. \cite{Wang2023CSCL} focus on continuous shape distribution at pixel level and propose Continuous Surface Correspondence Learning (CSCL) to enhance the global understanding of human body shape by shape embedding paradigm based on 2D-3D correspondence. 
These methods all use additional modal information (3D, shape, and contour, etc) to assist the model in learning. However, this inevitably introduces uncontrollable overhead and noise. As a result, a batch of methods that deeply explore the implicit information contained within the image itself have emerged. For example, Gu et al. \cite{Gu2022CAL} designed a clothes-based Adversarial Loss (CAL) to penalize the model's ability to predict clothes, mining features unrelated to clothes from the original RGB images. Zhao et al. \cite{Zhao2023imsgep} regard clothing change as a fine-grained domain/style transfer, and propose a joint Identity-aware Mixstyle and Graph-enhanced Prototype for cloth-changing person Re-ID. Gao et al. \cite{Gao2023SAVS} proposed a new semantic-aware attention and visual shielding network (SAVS), which shields the cues related to clothing appearance and only focuses on visual semantic information that is insensitive to viewpoint/pose changes. Yang et al. \cite{Yang2023AIM} designed a causality-based Auto Intervention Model (AIM) for cloth-changing person ReID, which captures clothing bias and identity cues separately, and simulates causal intervention by stripping clothing inference from identity representation learning. Guo et al. \cite{Guo2023SCNet} proposed a Semantic-aware Consistency Network (SCNet) to learn identity-related semantic features by proposing effective consistency constraints. Cui et al. \cite{Cui2023DCR-ReID} proposed Deep Component Reconstruction Re-ID (DCR-ReID), which realizes the controllable decoupling of clothes-independent features and clothes-related features. Yang et al. \cite{Yang2023ACID} focus on potential identity cues hidden in appearance and structural features, and propose an Auxiliary-free Competitive IDentification (ACID) model to perform precise identity identification without auxiliary data. Additionally, Wu et al. \cite{Wu2023CT-Net} proposed a two-stream hybrid Convolution Transformer Network (CT-Net), which combines CNN and Transformer in parallel in an end-to-end learning scheme.

Due to the large amount of redundant spatial information in visual images \cite{he2022mae}, the existing methods cannot decouple this information well, resulting in poor performance of the existing methods. Therefore, in this work, we will focus on several aspects (i.e., i. \emph{Control the spatial redundant information}, ii. \emph{Use more identity cues}, iii. \emph{reducing feature bias}) energetically explore robust representations of a person with different clothing.

\section{Multiple Information Prompt Learning Network}

\begin{figure*}
  \centering
  \includegraphics[width=\linewidth]{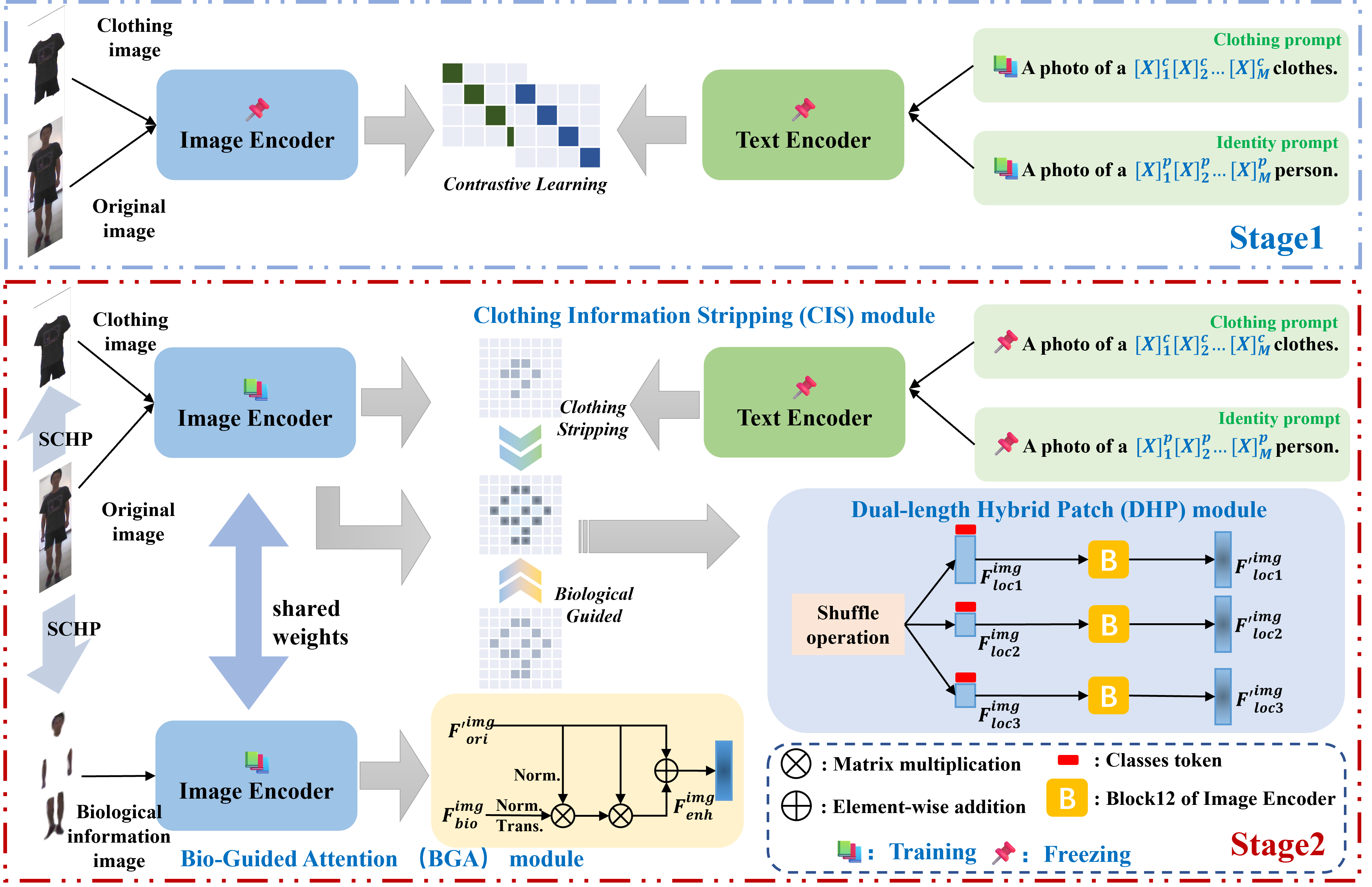}
  \caption{Pipeline of the proposed MIPL approach. It consists of the backbone, the CIS module, the BGA module, and the DHP module. It is a two-stage model with stage 1 and stage 2. In stage 1, the text prompts (Clothing prompt and Identity prompt) are optimized, and in stage 2, the image encoder is trained. 'SCHP' is a semantic analysis module to obtain the human semantic information. 'Norm.' and 'Trans.' denote the normalization and transpose operations, respectively. Other marks are marked in the lower right corner of pipeline.}
  \label{fig:framework-MIPL}
\end{figure*}

In this work, we develop a new MIPL algorithm for cloth-changing person ReID, which applies multiple aspects of information to co-prompts guided model learning. It can decouple redundant information in visual modes, make full use of easily captured pedestrian identity clues, reduce the influence of feature bias, and enable the model to learn the features of identity robustness. MIPL network is mainly composed of CIS module, BGA module and DHP module. The framework is shown in Figure \ref{fig:framework-MIPL}. Specifically, the model is divided into two stages. In the first stage, only the CIS module and backbone network participate in the training, freezing the parameters of the image and the text encoder, optimizing a set of learnable text prompt words for each identity and clothing; 
In the second stage, the BGA module and DHP module are added to freeze the text encoder and optimized the text prompt words, and fine-tune the image encoder. In addition, the clothing image of the CIS module and the biological information image of the BGA module are both obtained with the assistance of the human parsing model SCHP \cite{Li2022SCHP}, and the output of the backbone network is fed to the DHP module to learn the features of diverse coverage. It is important that they learn together in a unified framework. In the following sections, we will cover each module and the loss function separately.

\subsection{Clothing Information Stripping (CIS)}

In order to effectively decouple the visual redundant information in the visual modality, inspired by the two-stage prompt learning \cite{radford2021clip}, in MIPL, the image encoder is implemented as a Transformer architecture like Vit-B/16 \cite{dosovitskiy2021ViT} to generate image representations. The text encoder is implemented as a language Transformer architecture such as BERT \cite{DevlinCLT2019Bert} to generate text representations. Next, the image features and text features are normalized and linearly projected into the cross-modal embedding space for visual language contrastive learning. To be specific, we pre-train the learnable text prompt words of identity and clothing to supplement the text information, to establish an effective correspondence between visual representations and high-level language descriptions, and constrain the model to accurately locate the clothing area through the text description to decouple the clothing area from the non-clothing area. It reduces the influence of clothing information on CC-ReID task. Specifically, in the first training stage, A set of learnable prompt words are introduced, which are an identity-dependent text prompt ("A photo of a $[X]_1^p$ $[X]_2^p$$...$ $[X]_M^p$ person.") and a clothes-dependent text prompt ("A photo of a $[X]_1^c$ $[X]_2^c$$...$ $[X]_M^c$ clothes.") Where each $[x]_*^*$is a learnable text token with the same dimension as the embedded word,$p$ denotes the text token belonging to the identity dependency, $c$ denotes the text token belonging to the clothing dependency, and $M$ denotes the number of learnable text tokens. Then we use the text encoder and image encoder with frozen parameters to obtain the corresponding text features $F_{ori}^{text}$, $F_{clo}^{text}$ and image features $F_{ori}^{img}$, $F_{clo}^{img}$ (the encoder is pre-trained by CLIP \cite{radford2021clip}). A contrastive learning loss function is used to constrain the alignment between text features and image features. 

For the image-text contrastive loss is calculated as:
\begin{small}
\begin{equation}
\begin{aligned}
\mathcal{L}_{i2t}^{ID}(y_i) = -\log \frac{\exp \{s(V_{y_i}, T_{y_i})/\tau \}}{\sum_{a=1}^{B}{ \exp \{s(V_{y_i}, T_a)/\tau \}}}
\end{aligned}
\end{equation}
\label{eqi2tID}
\end{small}
Where $V$ is the image feature embedding, $T$ is the text feature embedding, $s(\cdot,\cdot)$ represents the inner product similarity calculation, $\tau$ is the temperature coefficient, and $B$ denotes the batch size.
Moreover, for the text-to-image contrastive loss, the text embedding $T$ may have multiple positives (there may exist multiple different images of a pedestrian in the same batch), so $\mathcal{L}_{t2i}$ is calculated as:
\begin{small}
\begin{equation}
\begin{aligned}
\mathcal{L}_{t2i}^{ID}(y_i) = \frac{-1} {|P(y_i)|} \sum_{p \in P(y_i)}{ \log \frac{\exp \{s(V_p, T_{y_i})/\tau \}}{\sum_{a=1}^{B}{ \exp \{s(V_a, T_{y_i})/\tau \}}}}
\end{aligned}
\end{equation}
\label{eqt2iID}
\end{small}
Among them, $P(y_i)$ is the set of indices of all positives for $T_{y_i}$ in the batch, and $|\cdot|$ is its cardinality. In addition, for $\mathcal{L}_{i2t}^{C}(y_c)$ and $\mathcal{L}_{t2i}^{C}(y_c)$ follow the above calculation.
In this way, a unique prompt is learned for different identities and clothes separately, providing precise guidance for the decoupling of clothing information from the original image. 

In the second training stage, we only optimize the image encoder and freeze the trained text prompt words and the parameters of the text encoder. The trained text features are used to align the clothing and body regions, and then the clothing stripping loss is designed to decouple the clothing information from the identity information in the image. 

Specifically, the image-to-text cross-entropy loss is calculated using the text embeddings trained in the first stage (including $N_i$ identity text embeddings and $N_c$ clothing text embeddings), and the image features are aligned by the text embeddings to guide the model to extract more accurate features, and the guide loss is calculated as:

\begin{small}
\begin{equation}
\begin{aligned}
\mathcal{L}_{Guide} &= \mathcal{L}_{i2tce}^{ID}(i) + \mathcal{L}_{i2tce}^{CLO}(c)\\
&=\sum_{m=1}^{N_i} -q_m {\log\frac{\exp \{s(V_i, T_m)\}}{\sum_{a=1}^{N}{ \exp \{s(V_i, T_a)\}}}}\\
& + \sum_{m=1}^{N_c} -q_m {\log\frac{\exp \{s(V_c, T_m)\}}{\sum_{a=1}^{N_c}{ \exp \{s(V_c, T_a)\}}}}
\end{aligned}
\end{equation}
\end{small}
For the clothing feature stripping operation, firstly, the consistent alignment operation is performed between the clothing mapping feature $F_{img2clo}^{img}$ and the clothing feature $F_{clo}^{img}$ by $\mathcal{L}_{sc}$. Through the knowledge mapping from the clothing feature, the identity information in the clothing feature is masked, so that the clothing mapping feature can be better aligned with the original feature. Then the feature decoupling loss $\mathcal{L}_{de}$ is calculated between the clothing mapping feature and the original feature, so that the original feature can include identifiable identity information other than clothing, and the calculation of $\mathcal{L}_{sc}$ and $\mathcal{L}_{de}$ are defined as follows:

\begin{small}
\begin{equation}
\begin{aligned}
\mathcal {L}_{sc} = \frac{1}{B} \sum_{i=1}^{B}{(F_{img2clo}^{img}-F_{clo}^{img})^2}
\end{aligned}
\end{equation}
\end{small}
\begin{small}
\begin{equation}
\begin{aligned}
\mathcal {L}_{de} = max\{0, COS(F_{ori}^{img}, F_{img2clo}^{img})\}
\end{aligned}
\end{equation}
\end{small}

Where $(\cdot)^2$ indicates the $L_2$ regularization and $COS(F_{ori}^{img}, F_{img2clo}^{img})$ is the cosine similarity between the original feature $F_{ori}^{img}$ and the clothing map feature $F_{img2clo}^{img}$.

For the clothing stripping loss $\mathcal{L}_{CS}$, the process is defined as follows:

\begin{small}
\begin{equation}
\begin{aligned}
\mathcal{L}_{CS} = \mathcal{L}_{Guide} + \mathcal{L}_{sc} + \mathcal{L}_{de}
\end{aligned}
\end{equation}
\end{small}

In addition, the spatial consistency loss is introduced to regularize the clothing mapping features (obtained by passing the original features through the last block of the image encoder) and clothing features into a unified feature hypersphere, so that the clothing features can be better aligned, to complete the feature stripping. It should be noted that the above feature operation details will be given in the loss function. 

\subsection{Bio-guided Attention (BGA)}

The existing work is not comprehensive enough to exploit the key information of pedestrians, and there is no explicit hint to learn comprehensive identity key features. Most of the work directly learns identity features implicitly from the original features or uses the head information for learning, so the mining of identity key information is not targeted and not comprehensive. In addition, we find that in the cloth-changing person ReID scenario, other biological information in the non-clothing area is relatively robust to identity information, such as arms, legs, and feet. (Usually shoes transform less frequently than clothes, so the shoes of pedestrians can help the model to extract identity information to a certain extent.) Therefore, to solve this problem, we propose the BGA module, which explicitly prompts the model with attention through unique biological key features. Specifically, in BGA module, the human body parsing model SCHP is used to obtain the biological key information masks corresponding to the original image, such as head, arms, left and right feet and legs. The biological information image is obtained by combining the mask image with the original image, and the obtained biological information image is input into the image encoder to obtain the biometric feature embedding $F_{bio}^{img}$. At the same time, we clone a new original feature named as ${F'}_{ori}^{img}$ for subsequent attention enhancement and knowledge distillation learning operations in the BGA module. The attention enhancement operation is performed through $F_{bio}^{img}$ and ${F'}_{ori}^{img}$ to explicitly emphasize the information of the biological key regions, and the information enhancement features $F_{enh}^{img}$ for the model to learn the prompts are obtained. The operation is defined by 

\begin{equation}
\begin{aligned}
\mathcal{M} = \mathcal{N}(F_{bio}^{img})^T \otimes \mathcal{N}({F'}_{ori}^{img}),
\end{aligned}
\end{equation}
\begin{equation}
\begin{aligned}
F_{enh}^{img} = \mathcal{M} \otimes {F'}_{ori}^{img} + {F'}_{ori}^{img},
\end{aligned}
\end{equation}

where $\mathcal{N}$ represents the normalization operation, $T$ represents the transpose operation, $\otimes$ represents matrix multiplication, and $\mathcal{M}$ indicates the biological key information mask.
In order to effectively use the information enhancement features, the biological guided loss is used to transfer the knowledge of biological key area information to the backbone network branches, which prompts the model to strengthen the learning of strong identity-related regions. 
The biological guided loss $\mathcal{L}_{BG}$ is calculated as:

\begin{small}
\begin{equation}
\begin{aligned}
\mathcal{L}_{BG} =&  \mathcal {D}_{KL}(p^{img} \| p^{bio}) + \mathcal {D}_{KL}(p^{bio} \| p^{img})\\
&= \sum_{i=1}^B p_i^{img} \log \frac {p_i^{img}}{p_i^{bio}}+\sum_{i=1}^B p_i^{bio} \log \frac {p_i^{bio}}{p_i^{img}}
\end{aligned}
\end{equation}
\end{small}
Where $p^{img}$ denotes output class probabilities of the backbone, $p^{bio}$ denotes output class probabilities of the BGA, Kullback-Leibler (KL) divergence \cite{zhang2018KL} is used to quantify the matching degree of the two output probabilities, and $\mathcal {D}_{KL}$ represents the KL distance. 


In this way, the backbone is endowed with the ability to extract the biological knowledge with strong identity correlation from the features, so as to improve the feature robustness of person reID in the cloth-changing scenario.

\subsection{Dual-length Hybrid Patch (DHP)}

In person ReID tasks, feature extraction is affected by objective factors such as pedestrian posture, occlusion, and shooting angle. Existing work mainly uses auxiliary models to overcome such feature biases, such as introducing gait models to learn pedestrian gait features, introducing edge detection models to learn pedestrian contours, etc. These methods are greatly limited by the performance of the pre-trained model and the quality of the transformed images. However, the image quality in the existing public dressing datasets is not high, which seriously affects the performance of such models. Therefore, inspired by ShuffleNet \cite{zhang2018shufflenet}, we propose the DHP module, which tries to fully explore the discriminative information with more diverse coverage from the features themselves and alleviate the impact of feature bias through special feature shuffling and grouping operations. Specifically, we take the original feature learned by prompting as the input of the DHP module, denoted as $[t^0,t^1,t^2,...,t^N]$, perform patch embedding random shuffling operation on the feature (except the category token [cls], i.e. $[t^1,t^2,...,t^N]$) to obtain the shuffled feature $[t^{s1},t^{s2},...,t^{sN}],s_i\in[1,N]$, and then, the shuffled features are truncated and divided into three groups of features with two lengths, and the shared category token $t^0$ is connected respectively. In this way, the local fine-grained features $F_{loc1}^{img}=[t^0,t^{s1},...,t^{sN/2}]$,$F_{loc2}^{img}=[t^0,t^{sN/2+1},...,t^{sN/4}]$, and $F_{loc3}^{img}=[t^0,t^{sN/4+1},...,t^{sN}]$ are obtained. Where $N$ denotes the number of patch embeddings. In addition, the local characteristics of fine-grained further from the last block of the image encoder for attention code embedded learning, get the local characteristics of fine-grained ${F'}_{loc1}^{img}$, ${F'}_{loc2}^{img}$, and ${F'}_{loc3}^{img}$. After shuffling and grouping, the dual-length hybrid patch embedding features cover several random patch embeddings from different body parts of the human body, and have dense and sparse coverage respectively, which endow the local features with the ability to recognize global information. In addition, the original feature $F_{ori}^{img}$ and the local features ${F'}_{loc1}^{img}$, ${F'}_{loc2}^{img}$, and ${F'}_{loc3}^{img}$ are concatenated as the final feature representation to balance the feature bias of the original features caused by objective factors such as pedestrian posture, occlusion, and shooting Angle.

\subsection{Loss Function}

For the optimization of MIPL network parameters, a two-stage training plan is implemented. 
\textbf{\\The first training stage.} In the first stage, we freeze the parameters of the image encoder and text encoder. And optimize the identity-dependent text prompt $[X]_m^p$ and cloth-dependent text prompt $[X]_m^c$ by contrastive learning, where $m\in[1, M]$, in preparation for the second stage of clothing decoupling. The contrastive learning loss for the first stage is defined by

\begin{small}
\begin{equation}
\begin{aligned}
\mathcal{L}_{Stage1} = 
 \mathcal{L}_{i2t}^{ID}(y_i) + \mathcal{L}_{i2t}^{C}(y_c)
+\mathcal{L}_{t2i}^{ID}(y_i) + \mathcal{L}_{t2i}^{C}(y_c)
\end{aligned}
\end{equation}
\end{small}
which includes the image-text contrastive loss $\mathcal{L}_{{i2t}}$ and the text-image contrastive loss $\mathcal{L}_{{t2i}}$, where $ID$ and $C$ represent identity and clothing dependency respectively, and  $y_i$ represents identity labels, and $y_c$ represents clothing labels. 


\textbf{\\The second training stage.} In this stage, both the text prompt and the text encoder are frozen and only the image encoder is optimized, we follow previous ReID work \cite{he2021transreid,li2023clip-reid,Zhou22OS-Net} to calculate cross-entropy loss and triplet loss to optimize the image encoder. 
For the cross-entropy loss $\mathcal{L}_{ce}$ and the triplet loss $\mathcal{L}_{tri}$, they are calculated as follows:

\begin{small}
\begin{equation}
\begin{aligned}
\mathcal{L}_{ce} = \frac{1}{B} \sum_{i=1}^{B} - \log p(x\mid y)
\end{aligned}
\end{equation}
\end{small}
\begin{small}
\begin{equation}
\begin{aligned}
\mathcal{L}_{tri} =  \frac{1}{B} \sum_{i=1}^{B} \max \{m + d_p - d_n, 0\}
\end{aligned}
\end{equation}
\end{small}
Where $p(x\mid y)$ is the predicted probability that the sample $x$ belongs to the ground truth $y$. $d_p$ and $d_n$ are the feature distances of positive pair and negative pair, and $m$ is the margin of $\mathcal{L}_{tri}$. 

In addition, in order to strip the clothing interference information, we design the clothing stripping loss to guide the model to decouple and strip the clothing information. In order to effectively use biological information to enhance features, the biological guided loss is used to prompt the model to learn biological knowledge with a strong identity correlation. 
Thus, the loss function in the second stage is defined by

\begin{small}
\begin{equation}
\begin{aligned}
\mathcal{L}_{Stage2} = \mathcal{L}_{ce} + \mathcal{L}_{tri} + \mathcal{L}_{CS} + \mathcal{L}_{BG},
\end{aligned}
\end{equation}
\end{small}

\section{Experiments And Discussion}

\begin{table*}
\fontsize{9}{9}\selectfont
\caption{Performance evaluation and comparison on five public cloth-changing datasets, where the bold values indicate the best performance in each column, and underlined values indicate the optimal performance of the existing methods. The MIPL* denotes the scheme of overlapping patches is adopted.}
\renewcommand{\arraystretch}{1.2}
\tabcolsep=0.26cm
\vspace{-0.5em}
\begin{center}
\begin{tabular}{c|cc|cc|cc|cc|cc}
\hline
\multirow{3}{*}{Methods} &\multicolumn{10}{c} {Datasets} \\
\cline{2-11}
& \multicolumn{2}{c|} {PRCC} & \multicolumn{2}{c|} {LTCC} & \multicolumn{2}{c|} {Celeb-reID} & \multicolumn{2}{c|} {Celeb-reID-light} & \multicolumn{2}{c} {CSCC}   \\
\cline{2-11}
                                        &mAP  &rank-1 &mAP&rank-1 &mAP&rank-1 &mAP&rank-1 &mAP&rank-1\\
\hline
ResNet50 \cite{he2016ResNet}             &8.1  &19.6    &8.4 &20.7     &5.8 &43.3     &6.0 &10.3 &13.1 &32.0   \\
PCB \cite{sun2018PCB}                    &-    &22.9    &8.8 &21.9     &8.2 &37.1     &- &- &15.5 &37.6   \\
MGN \cite{wang2018MGN}                   &-    &25.9    &10.1 &24.2     &10.2 &48.6     &13.9 &21.5  &- &-    \\
ViT-B/16 \cite{dosovitskiy2021ViT}       &46.4 &46.3    &28.6 &69.5     &- &-     &17.1 &30.2  &- &-    \\
\hline
FSAM \cite{Hong2021FSAM}                 &- &54.5 &16.2 &38.5   &- &-    &- &-  &- &-    \\
3DSL \cite{Chen20213DSL}                 &- &51.3 &14.8 &31.2   &- &-    &- &-  &- &-    \\
MAC-DIM \cite{Chen2022MAC-DIM}           &- &48.8 &13.0 &29.9    &- &-    &- &-  &- &-   \\
LaST \cite{shu2022LaST}                  &54.7 &57.5    &- &-     &11.8 &54.4     &16.3 &29.0  &- &-  \\
GI-ReID \cite{Jin2022GI-reid}            &- &37.6    &14.2 &28.9    &- &-    &- &-  &- &- \\
CAL \cite{Gu2022CAL}                     &55.8 &55.2    &18.0 &40.1     &- &-    &- &-   &- &- \\
MVSE \cite{Gao2022MVSE}                  &52.5 &47.4    &33.0 &70.5     &19.2 &64.5     &- &-  &- &-  \\
UCAD \cite{yan2022UCAD}                                     &- &45.3    &15.1 &32.5     &- &-    &- &- &25.9 &53.8    \\
Pos-Neg \cite{Liu23Dual}                                     &65.8 &54.9    &14.1 &36.2     &- &-    &- &- &- &-    \\
DCR-ReID \cite{Cui2023DCR-ReID}          &57.2 &57.4    &20.4 &41.1     &- &-    &- &-  &- &-  \\
AIM \cite{Yang2023AIM}                   &58.3 &57.9    &19.1 &40.6     &- &-    &- &-  &- &-  \\
CCFA \cite{Han2023CCFA}                  &58.4 &61.2    &22.1 &45.3     &- &-    &- &-   &- &- \\
SAVS \cite{Gao2023SAVS}                  &57.6 &\underline{69.4}    &32.5 &71.2     &\underline{21.3} &\underline{65.9}     &- &-  &- &-  \\
IMS+GEP \cite{Zhao2023imsgep}            &65.8 &57.3    &18.2 &43.4     &- &-    &- &-   &- &- \\
PGAL \cite{Liu2023PGAL}                  &58.9 &59.7    &27.7 &62.5     &15.3 &60.9     &\underline{23.3} &\underline{40.4}  &- &-  \\
CT-Net \cite{Wu2023CT-Net}               &61.3 &48.9    &\underline{37.5} &\underline{72.4}     &13.7 &60.2     &- &- &- &-   \\
ACID \cite{Yang2023ACID}                 &\underline{66.1} &55.4    &14.5 &29.1     &11.4 &52.5     &15.8 &27.9 &- &-   \\
SCNet \cite{Guo2023SCNet}                &59.9 &61.3    &25.5 &47.5     &- &-    &- &-   &- &-  \\
CSCL \cite{Wang2023CSCL}                 &64.5 &64.2    &34.1 &69.7     &- &-    &- &-  &- &-  \\
DLAW \cite{Liu23Dual}                 &57.1 &56.2    &35.6 &58.0     &- &-    &- &-  &- &-  \\
MBUNet \cite{Zhang2023Multi}                 &65.2 &68.7    &15.0 &40.3     &12.8 &55.5    &21.5 &35.5  &- &-  \\
\hline
Baseline    
&56.7 &61.5    
&33.5 &72.0     
&32.3 &72.1    
&45.2 &63.1     
&46.6 &86.9    \\
\textbf{MIPL(Ours)}
&64.8 &69.2    
&\textbf{38.1} &\textbf{74.8}     
&\textbf{33.2} &\textbf{73.3}    
&\textbf{47.4} &\textbf{66.0}   
&\textbf{47.1} &\textbf{88.1}    \\
\textbf{MIPL*(Ours)}    
&\textbf{67.0} &\textbf{71.0}    
&\textbf{38.8} &\textbf{75.1}     
&\textbf{37.0} &\textbf{76.2}   
&\textbf{50.6} &\textbf{69.8}   
&\textbf{46.4} &\textbf{88.1}    \\
\hline
\end{tabular}
\end{center}
\vspace{-1.0em}
\label{Performance}
\end{table*}

To evaluate the performance of the MIPL method, we performed experiments using five public cloth-changing person ReID datasets: PRCC \cite{yang2021PRCC}, LTCC \cite{qian2020LTCC}, Celeb-reID \cite{huang2020Celeb-reID}, Celeb-reID-light \cite{huang2019Celeb-reID-light}, and CSCC\cite{yan2022UCAD}. 
The remainder of this section is organized as follows: 1) five public cloth-changing person ReID datasets are introduced, 2) the competing methods used in our experiments are listed, 3) the implementation details are described, 4) the performance evaluations and comparisons based on these five public datasets are described, and 5) convergence analysis.

\subsection{Datasets}
Our experiments utilized five datasets, namely, PRCC\cite{yang2021PRCC}, LTCC\cite{qian2020LTCC}, Celeb-reID\cite{huang2020Celeb-reID}, Celeb-reID-light\cite{huang2019Celeb-reID-light}, and CSCC\cite{yan2022UCAD}. 
The PRCC, LTCC and CSCC datasets were assembled using images captured by real surveillance cameras while the Celeb-reID and Celeb-reID-light datasets was sourced from the internet. Note that for privacy reasons, all faces in the CSCC dataset are masked.

\subsection{Competitors}

The task of CC-ReID is a new and challenging topic that has also aroused the interest of researchers in related fields in recent years. In our experiments, the latest and popular references were utilized as our competitors, including FSAM (CVPR 2021) \cite{Hong2021FSAM}, 3DSL (CVPR 2021) \cite{Chen20213DSL}, MAC-DIM (TMM 2021) \cite{Chen2022MAC-DIM}, LaST (TCSVT 2021) \cite{shu2022LaST}, GI-ReID (CVPR 2022) \cite{Jin2022GI-reid}, CAL (CVPR 2022) \cite{Gu2022CAL}, MVSE (ACM MM 2022) \cite{Gao2022MVSE}, UCAD (IJCAI 2022) \cite{yan2022UCAD}, Pos-Neg (TIP 2022) \cite{Jia2022Complementary}, DCR-ReID (TCSVT 2023) \cite{Cui2023DCR-ReID}, AIM (CVPR 2023) \cite{Yang2023AIM}, CCFA (CVPR 2023) \cite{Han2023CCFA}, SAVS (TNNLS 2023) \cite{Gao2023SAVS}, IMS+GEP (TMM 2023) \cite{Zhao2023imsgep}, PGAL (TMM 2023) \cite{Liu2023PGAL}, CT-Net (TMM 2023) \cite{Wu2023CT-Net}, ACID (TIP 2023) \cite{Yang2023ACID}, SCNet (ACM MM 2023) \cite{Guo2023SCNet}, CSCL (ACM MM 2023) \cite{Wang2023CSCL}, DLAW (TIP 2023) \cite{Liu23Dual}, and MBUNet (TIP 2023) \cite{Zhang2023Multi}. Additionally, in the CC-ReID task, traditional person ReID algorithms, such as ResNet50 (CVPR 2016) \cite{he2016ResNet}, ViT-B/16 (ICLR 2021) \cite{dosovitskiy2021ViT}, PCB (ECCV 2018) \cite{sun2018PCB}, and MGN (ACM MM 2018) \cite{wang2018MGN}, are often employed. In our experiments, we also compared MIPL with them. In addition, the introduction of non-RGB modal information can provide richer information to the model. For example, 3DSL, FSAM, MAC-DIM, GI-ReID, DCR-ReID, and PGAL all use different modal information. It is worth noting that MIPL is the first method to introduce text information as a prompt into the task of CC-ReID. More information about these competitors can be obtained in the related work section.

\subsection{Implementation Details}

We adopt the image encoder and text encoder of CLIP \cite{radford2021clip} as the backbone, and for the image encoder, we adopt ViT-B/16 with 12 transformer layers and the hidden size is 768 dimensions. The dimension of the image feature vector is reduced from 768 to 512 by a linear layer to match the dimension of the text feature vector output by the text encoder.
In training stage 1, we adopt Adam optimizer with learning rate initialized to $3.5e^{-4}$ and decay by a cosine schedule. The batch size is set to 64 without using any augmentation methods, and only the learnable text tokens are optimized. The number of learnable text tokens is set to 4.
In training stage 2, the minibatch size was set to 64. It contained 16 randomly selected pedestrian identities with 4 images per identity, and the input person images were resized to $256 \times 128$. Each image is augmented by random erasing. The Adam optimizer is also used to train the image encoder. The model was trained for 120 epochs. We first warm up the model for 10 epochs with a linearly growing learning rate from $5e^{-7}$ to $5e^{-6}$. Then, it is decreased by a factor of 0.1 at the 30th and 50th epochs. The temperature coefficient $\tau$ is set to 1.
Finally, the rank-1, and mean average precision (mAP) are often utilized as the evaluation metrics in person ReID tasks \cite{Ye2022AGW,Gao2022MVSE}, thus, we also strictly follow these metrics in our experiments. 

\subsection{Performance Evaluations and Comparisons}

In this section, we first evaluate the performance of the MIPL algorithm on five public cloth-changing person ReID datasets, and then compare it with the above competitors. Note that the original papers of GI-ReID \cite{Jin2022GI-reid}, AIM \cite{Yang2023AIM}, CCFA \cite{Han2023CCFA}, PGAL \cite{Liu2023PGAL}, and MBUNet \cite{Zhang2023Multi} reported multiple sets of results for the same method with different variable controls, for example, the GI-ReID method reports the results of multiple baselines, the AIM method reports the results obtained with multiple input sizes, the CCFA method reports the results before and after adding feature enhancement, the PGAL method reports the results of adding multi-granularity perception, and the MBUNet method reports the results of adding mAP optimization, in such cases, we choose the highest results reported by them to compare with them. In addition, The MIPL* denotes the Overlapping Patches \cite{he2021transreid} is adopted. The results are shown in Table \ref{Performance}. From these results, we obtain the following observations:

1) No matter which method is compared with, the MIPL algorithm achieves the best performance on LTCC, Celeb-reID, Celeb-reed-Light and CSCC datasets, and significantly improves the mAP and rank-1 compared with the existing algorithms. For the PRCC dataset, MIPL still has comparable performance with existing algorithms. 
For example, the mAP and rank-1 accuracy of MIPL on the LTCC dataset are 38.1\% and 74.8\%, respectively, while the corresponding performance of Baseline is 33.5\% and 72.0\%, respectively. The improvement level can reach 4.6\% (mAP) and 2.8\% (rank-1), respectively. When using the PRCC dataset, the mAP and rank-1 of MIPL are 64.8\% and 69.2\%, respectively, while the corresponding performance of Baseline is 56.7\% and 61.5\%, respectively, and the corresponding improvement reaches 8.1\% (mAP) and 7.7\% (Rank-1).
In addition, when the overlapping patch enhancement strategy is adopted, the performance of MIPL is further improved, and the mAP and Rank-1 accuracy of MIPL* reach 67.0\% and 71.0\%, respectively. MIPL* algorithm comprehensively exceeds the performance of the existing optimal algorithms and is significantly better than Baseline. Through the two-stage training strategy, the model can more accurately decouple the clothing interference factors. Through the joint optimization of CIS module, BGA module and DHP module embedded into the baseline, the model can simultaneously obtain multiple aspects of information tips and extract more effective identity robust features. Therefore, MIPL shows good generalization ability in experiments, and these experimental results demonstrate the effectiveness and robustness of the proposed method. 

 2) When compared with specifically designed clothing-changing person ReID methods, PGAL achieved the second best performance on Celeb-reID-light dataset, with mAP and Rank-1 of 23.3\% and 40.4\%, respectively. The corresponding performance of MIPL is improved by 24.1\% (mAP) and 25.6\% (Rank-1). On the LTCC dataset, the mAP and Rank-1 of CT-Net are 37.5\% and 72.4\%, respectively, while the mAP and Rank-1 of MIPL are 38.1\% and 74.8\%, which are increased by 0.6\% and 2.4\%, respectively. When using the PRCC dataset, the mAP/Rank-1 of ACID and MIPL are 66.1\%/55.4\% and 64.8\%/69.2\%, respectively. The mAP accuracy of MIPL is 1.3\% lower than that of ACID, but the rank-1 accuracy of MIPL is 13.8\% higher than that of ACID. On the Celeb-reID dataset, the mAP and rank-1 of the MBUNet algorithm are 12.8\% and 55.5\%, respectively. The corresponding performance of the MIPL algorithm shows an improvement of 20.4\% in mAP and 17.8\% in Rank-1. When LTCC dataset is used, the mAP/Rank-1 of ACID and MIPL are 14.5\%/29.1\% and 38.1\%/74.8\%, respectively, and the performance of MIPL method is significantly better than that of ACID method. These methods specifically designed for the cloth-changing problem mainly introduce additional human keypoint information from multimodal information or model human shape to extract low-level features to avoid the interference of clothing information. Or implicitly learn from the overall features, without prompting guidance for the interference factors of the task pain points, making the model difficult to learn and lack of targeted high-level semantic information extraction. MIPL extracts semantic information strongly related to identity from high-level features through prompt learning, which has good robustness and generalization for the problem of cloth-changing. Overall, MIPL consistently outperforms all these SOTA methods on several publicly available datasets, demonstrating the effectiveness of MIPL.

3) For ResNet50, ViT-B/16, PCB and MGN, the first two models are widely used in computer vision tasks, and they are also often evaluated on person ReID tasks. The latter is some traditional person ReID methods that mainly learn pedestrian features based on the appearance of pedestrians' clothes. Although these methods have excellent performance in related tasks, they cannot perform well when they are directly used to deal with the person ReID task of cloth-changing. MIPL has an absolute advantage over these methods on the cloth-changing person ReID task.

\section{Ablation Study}

An ablation study was performed using the MIPL model to analyze the contribution of each component. In this investigation, three aspects were considered: 1) the effectiveness of the CIS module, 2) the advantages of the BGA module, and 3) the benefits of the DHP module. In the following, we discuss these three aspects separately.

\subsection{Effectiveness of the CIS Module}

\begin{table}
\fontsize{10}{10}\selectfont
\caption{Effectiveness of the CIS Module.}
\renewcommand{\arraystretch}{1.5}
\vspace{-1.0em}
\begin{center}
\begin{tabular}{c|cc|cc}
\hline
\multirow{3}{*}{Methods} &\multicolumn{4}{c} {Datasets} \\
\cline{2-5}
& \multicolumn{2}{c|} {PRCC} &\multicolumn{2}{c} {LTCC} \\
\cline{2-5}
&mAP &rank-1 &mAP &rank-1\\
\hline
Baseline                          &56.7 &61.5 &33.5 &72.0 \\
+CIS [w/o clo.prompts]           &61.3  &64.7  &33.6  &70.0   \\
+CIS [w/ clo.prompts]            &\textbf{63.3}  &\textbf{66.0}  &\textbf{38.1}  &\textbf{73.3}   \\
\hline
\end{tabular}
\end{center}
\vspace{-1.5em}
\label{table_CIS}
\end{table}

In many existing person ReID methods, human contour, key points, or gait features are usually used to model the associations unrelated to clothing and resist the interference of clothing information. In this section, we evaluate the effectiveness of the CIS module, considering the importance of targeting the clothing region for decoupling via text prompts. Since MIPL employs CLIP's image encoder and text encoder as the backbone, it is also used as a baseline in our experiments. Since the clothing region image is used in the CIS module to strip the clothing region in the original image, we analyze the effectiveness of the CIS module when the clothing information is decoupled without clothing text prompts and when the clothing information is decoupled using text prompts. The results are shown in Table \ref{table_CIS}, where '+CIS [w/o clo.prompts]' means that the clothing information is stripped without using the clothing text prompts, '+CIS [w/ clo.prompts]' means that the clothing information is stripped using the clothing text prompts, and from it, we can obtain the following observations:

When the operation of stripping clothing area information is directly embedded into the baseline, the performance of the model can be improved to a certain extent, but it is unstable for the scheme that does not use text information for prompting. For example, when using the PRCC dataset, The mAP/rank-1 of baseline and '+CIS [w/o clo.prompts]' are 56.7\%/61.5\% and 61.3\%/64.7\%, respectively, and the improvement is 4.6\%/3.2\%. However, when the LTCC dataset is selected, the mAP of '+CIS [w/o clo.prompts]' is increased by 0.1\% compared with the baseline, but the rank-1 is reduced by 2.0\%. The reason is that there are more occlusion and posture changes in the LTCC dataset, which will affect the stripping of clothing areas to a certain extent. When clothing alignment is aided by clothing text prompts, CIS consistently achieves the best performance regardless of the dataset used. 
For example, when using the LTCC dataset, the mAP/rank-1 accuracy of '+ CIS [w/clo prompts]' and baseline are 38.1\%/73.3\% and 33.5\%/72.0\%, respectively, and its improvement reaches 4.6\%/1.3\%. In addition, '+CIS [w/ clo.prompts]' is further improved by 4.6\% (mAP) /3.3\% (rank-1) compared to '+CIS [w/o clo.prompts]'. Similarly, when selecting the PRCC dataset, '+CIS [w/ clo.prompts]' further improves mAP/rank-1 by 2.0\%/1.3\% over '+CIS [w/o clo.prompts]'. These results prove the effectiveness of the CIS module, and the strategy of clothing text prompts can effectively improve the module's accurate distinction and location of clothing area information so that the clothing texture information unrelated to identity can be more accurately stripped, so as to effectively and reasonably reduce the interference of clothing texture information.

\subsection{Advantages of the BGA Module}

\begin{table}
\fontsize{10}{10}\selectfont
\caption{Advantages of the BGA Module.}
\renewcommand{\arraystretch}{1.5}
\vspace{-1.0em}
\begin{center}
\begin{tabular}{c|cc|cc}
\hline
\multirow{3}{*}{Methods} &\multicolumn{4}{c} {Datasets} \\
\cline{2-5}
& \multicolumn{2}{c|} {PRCC} &\multicolumn{2}{c} {LTCC} \\
\cline{2-5}
&mAP &rank-1 &mAP &rank-1\\
\hline
Baseline                        &56.7 &61.5 &33.5 &72.0 \\
+BGA [only head]                &\textbf{63.6}  &67.3  &37.2  &72.8   \\
+BGA [bio. info.]           &63.4  &\textbf{68.4}  &\textbf{37.7}  &\textbf{73.8}   \\
\hline
\end{tabular}
\end{center}
\vspace{-1.5em}
\label{table_BGA}
\end{table}

In this section, we assess the advantages of the BGA module. Similarly, MIPL's image encoder and text encoder adopting CLIP are used as baselines in the experiments. However, in the BGA module, different images are used to serve as guidance images for the BGA module, including head-only images and biological information images, which are then used to generate guidance masks. The results are shown in Table \ref{table_BGA}, where '+BGA [only head]' and '+BGA [bio. info.]' represent combining the BGA module with the baseline using head images and biological information images, respectively. From the results, it can be observed that when the BGA module is embedded in the baseline, the performance can be improved regardless of which type of image is chosen as the guide. For example, when using the PRCC dataset, the mAP/rank-1 accuracy of '+BGA [only head]' and the baseline is 63.6\%/67.3\% and 56.7\%/61.5\%, respectively, resulting in an improvement of 6.9\% (mAP) and 5.8\% (rank-1). Similarly, when using the LTCC dataset, '+BGA [only head]' achieves an improvement of 3.7\% (mAP) and 0.8\% (rank-1) over the baseline, respectively. Therefore, these results suggest that the BGA module is very effective and useful for guiding the model to learn identity robust information. Moreover, the performance is further improved when using biological information images with more identity-robust information applied to the generation of the guided mask. For example, compared to '+BGA [only head]', '+BGA [bio. info.]' achieves a further improvement of 0.5\% and 1.0\% in mAP/rank-1 accuracy on the LTCC dataset, respectively, and similarly, a further improvement of 1.1\% in rank-1 accuracy on the PRCC dataset. Using biological information images containing more identity-related information to generate guidance masks can enable the model to mine the key information of identity in a targeted and more comprehensive way, reduce the influence of interference factors such as clothing and background in the image on the strong identity-related information, and guide the model to explore more comprehensive identity robust information.

\subsection{Benefits of the DHP Module}

\begin{table}
\fontsize{10}{10}\selectfont
\caption{Benefits of the CIS, BGA, and DHP modules.}
\renewcommand{\arraystretch}{1.5}
\vspace{-1.0em}
\begin{center}
\begin{tabular}{c|cc|cc}
\hline
\multirow{3}{*}{Methods} &\multicolumn{4}{c} {Datasets} \\
\cline{2-5}
& \multicolumn{2}{c|} {PRCC} &\multicolumn{2}{c} {LTCC} \\
\cline{2-5}
&mAP &rank-1 &mAP &rank-1\\
\hline
Baseline         &56.7  &61.5  &33.5  &72.0  \\
+CIS             &63.3  &66.0  &38.1  &73.3  \\
+BGA             &63.4  &68.4  &37.7  &73.8  \\
+DHP             &62.2  &65.0  &37.7  &74.1  \\
+CIS+BGA         &63.7  &67.4  &\textbf{38.2}  &74.3  \\
+CIS+BGA+DHP     &\textbf{64.8}  &\textbf{69.2}  &38.1  &\textbf{74.8}  \\
\hline
\end{tabular}
\end{center}
\vspace{-1.5em}
\label{collaborative_modules_Table}
\end{table}

We next verify the benefits of DHP module by joint learning schemes of each module, in our experiments, CIS module, BGA module and DHP module are gradually embedded into the baseline, and then CIS module, BGA module and DHP module jointly optimize the model, and the results are shown in Table \ref{collaborative_modules_Table}. Note that in the table, CLIP's image encoder and text encoder are treated as baselines in the experiments. In addition, when the CIS module that performs the clothing information stripping strategy is embedded into the baseline, it is called '+CIS'. When adding the bio-guided BGA module to the baseline, it is named '+BGA'. When the DHP module is embedded into the baseline using the dual-length hybrid patching strategy, it is named '+DHP'. When the CIS module and BGA module are jointly embedded into the baseline, it is named '+CIS+BGA'. Finally, when we further add the DHP module to the '+CIS+BGA' module, we name it '+CIS+BGA+DHP'. We observe that as each module is gradually embedded into the baseline, their combined performance yields a steady boost (with the exception of mAP for LTCC) and that the individual modules are complementary to each other and they reinforce each other. For example, when using the PRCC dataset, the rank-1 accuracy of the baseline, '+CIS', '+CIS+BGA', and '+CIS+BGA+DHP' are 61.5\%, 66.0\%, 67.4\%, and 69.2\%, respectively, and their performance gradually improves with the combination of modules. In addition, compared with the baseline and '+CIS+BGA', the rank-1 accuracy improvement of '+CIS+BGA+DHP' reaches 7.1\% and 1.8\%, respectively. Similarly, on the LTCC dataset, '+CIS+BGA+DHP' achieves 2.8\% and 0.5\% improvement in rank-1 accuracy over the baseline and '+CIS+BGA', respectively. Therefore, these results demonstrate the benefits of the DHP module, which can alleviate the feature bias caused by pedestrian pose, occlusion, shooting viewpoint and other factors to some extent by fully exploring the identity discriminant information with more diverse coverage.

\subsection {Visualization Results}

\begin{figure}
\begin{minipage}{1\linewidth}
\includegraphics[width=\linewidth]{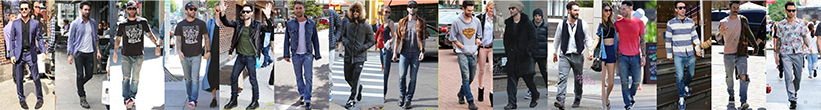}
\centering{(a) 15 images of the same person wearing different clothes} 
\end{minipage}
\\
\begin{minipage}{0.45\linewidth}
\centering{\includegraphics[width=\linewidth]{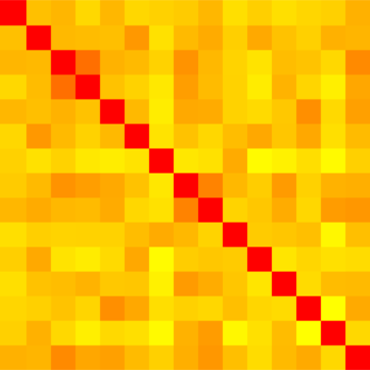}}
\centering{(b) similarity map with the Baseline}
\end{minipage}
\hfill
\begin{minipage}{0.45\linewidth}
\centering{\includegraphics[width=\linewidth]{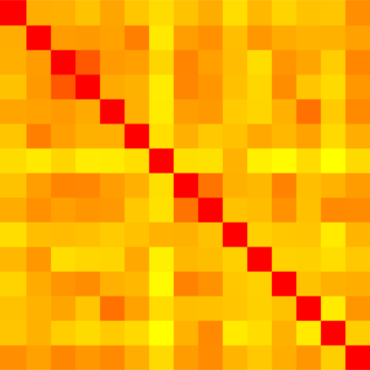}}
\centering{(c) similarity map with the MIPL}
\end{minipage}
\vspace{1em}

\begin{minipage}{1\linewidth}
\includegraphics[width=\linewidth]{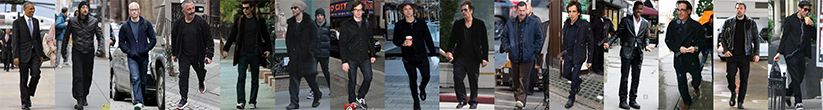}
\centering{(d) 15 images of different people wearing similar clothes}
\end{minipage}
\\
\begin{minipage}{0.45\linewidth}
\centering{\includegraphics[width=\linewidth]{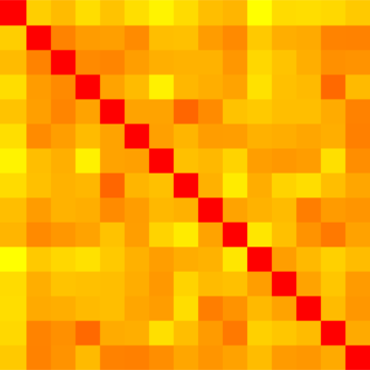}}
\centering{(e) similarity map with the Baseline}
\end{minipage}
\hfill
\begin{minipage}{0.45\linewidth}
\centering{\includegraphics[width=\linewidth]{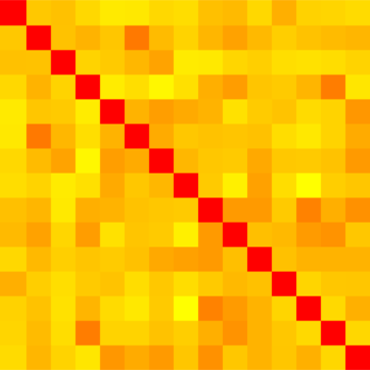}}
\centering{(f) similarity map with the MIPL} 
\end{minipage} 

\caption{Similarity matrices of the baseline and MIPL. Cosine similarity is used to calculate the distance between any two images. The color of each square indicates the similarity degree between these two images indicated by the horizontal and vertical coordinates. The red and yellow colors represent the most similar pairs and least similar pairs, respectively.} 
\label{fig:Person_Similarity_matrices}
\end{figure}

\begin{figure*}[ht]
\begin{center}
\includegraphics[width=\linewidth]{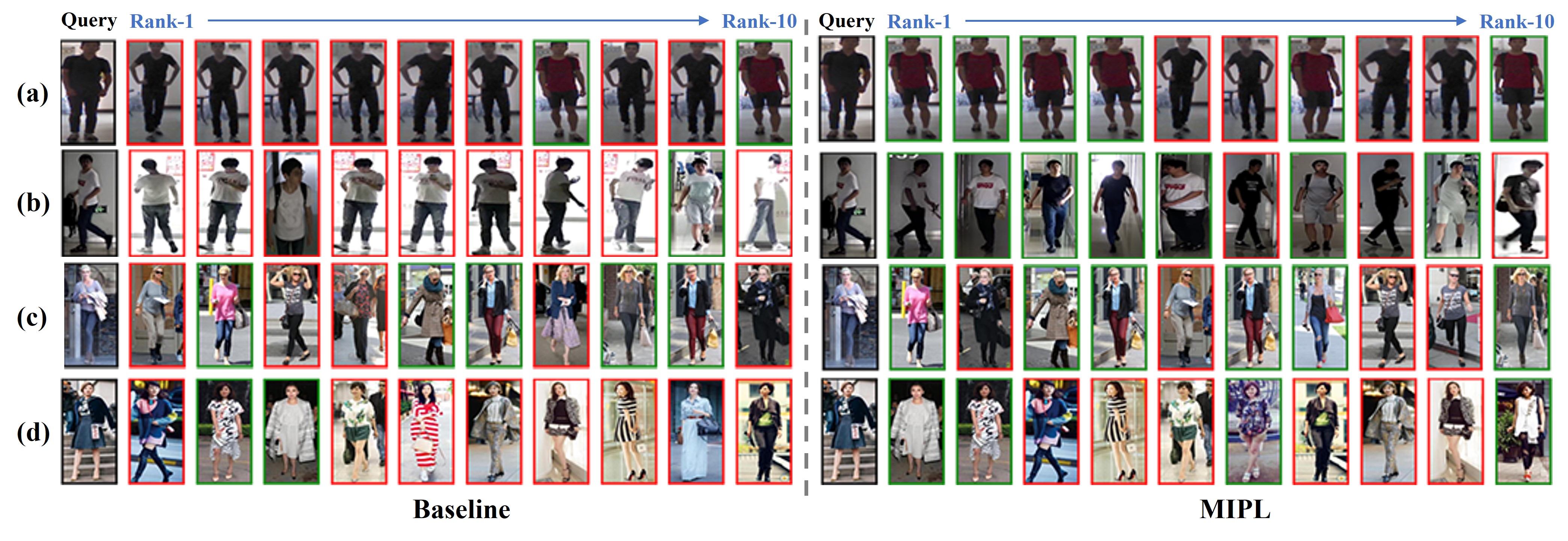}
\caption{Top-10 ranking results of MIPL and the baseline where different queries with different cases, such as front(a), side(b), and outdoor background clutter(c-d), are utilized. Green boxes indicate correct results and red boxes represent incorrect results.} 
\label{fig:Top_10_ranking_results}
\end{center}
\end{figure*}

To further demonstrate the effectiveness and robustness of MIPL, we visualize some experimental results while considering three aspects: 1) visualization of the attention maps, 2) visualization of the similarity map, and 3) qualitative visualization of the retrieval results. In what follows, we discuss these three aspects separately and the results are shown in Figures \ref{fig:grad-cam_MIPL}, \ref{fig:Person_Similarity_matrices}, and \ref{fig:Top_10_ranking_results}, where we make the following observations:

\begin{figure}
  \centering
  \includegraphics[width=\linewidth]{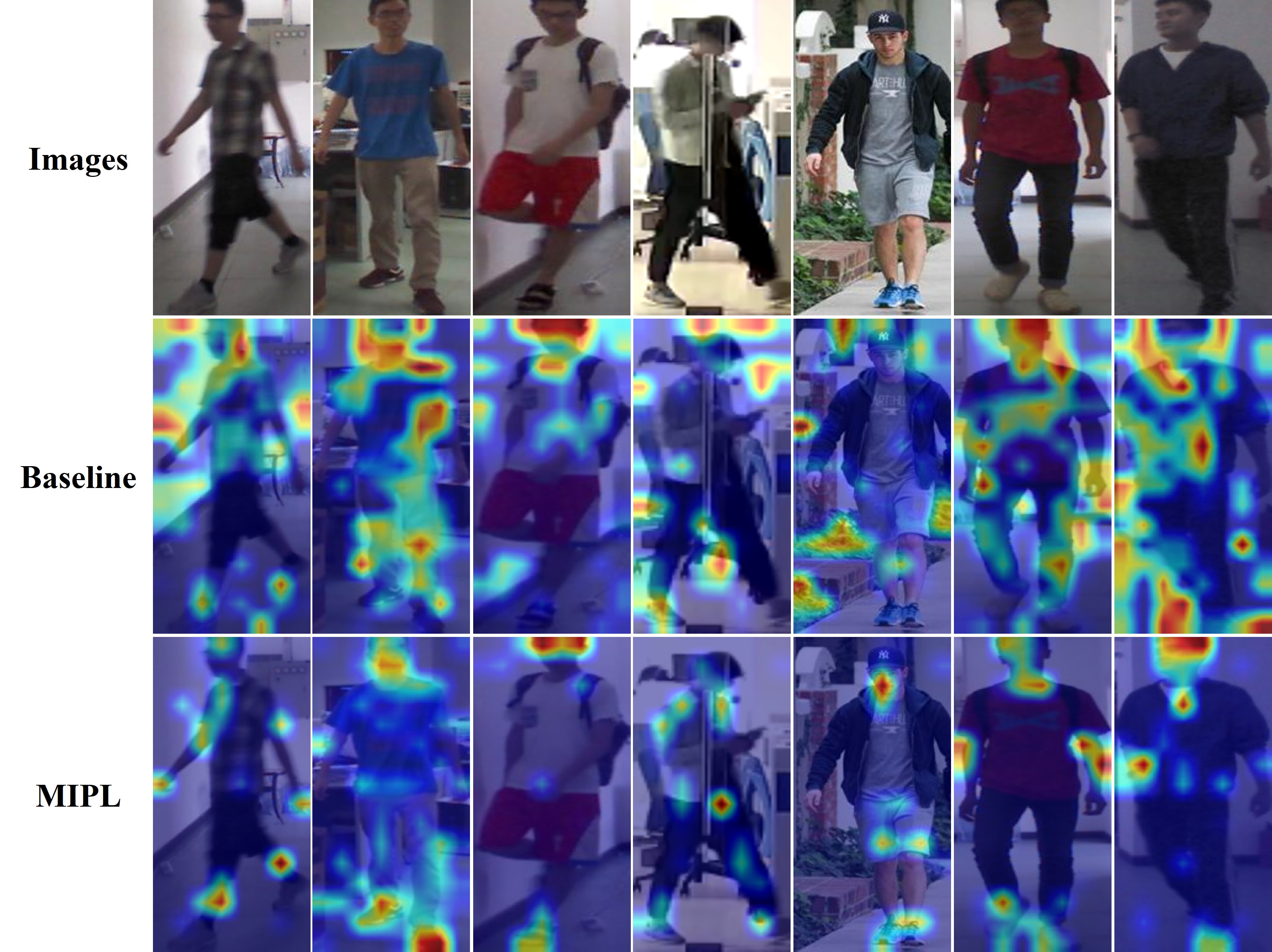}
  \caption{Visualization of attention maps. The first row, second row, and third row indicate the original images, attention maps of the baseline, and attention maps of MIPL, respectively. Note that the brighter the pixels, the more attention the model pays, and the identity of each column belongs to the same person.}
  \label{fig:grad-cam_MIPL}
\end{figure}

1) To better understand the working principle of different modules and to further illustrate which cues are more focused, we used the Grad-CAM \cite{selvaraju2017gradcam} method to visualize and display the intermediate activation feature maps of the baseline and MIPL in Figure \ref{fig:grad-cam_MIPL}. We observe that the activation feature maps of the baseline model mainly focus on global context information. In the process of feature extraction, more interference information (such as background and clothing texture) is introduced. However, the discrimination, clothing independence and generalization of these features are insufficient. In contrast, the activation feature map of MIPL pays more attention to biological information regions strongly related to pedestrian identity, and pays little attention to places containing clothing texture information, background and other interference factors. In addition, MIPL can also alleviate the interference caused by occlusion situations to a certain extent, and accurately focus the attention on the human body area. Therefore, these experiments further demonstrate the effectiveness and superiority of the proposed method.




2) To intuitively illustrate the effectiveness of MIPL from another perspective, we calculate the feature similarity between any two different images. Specifically, we selected 15 images of the same pedestrian wearing different clothes in the experiment and then used the baseline model to extract features from each image. Moreover, we calculated the cosine similarity between any two images based on the extracted features. We repeated the above operation in pairs for all 15 images and visualized their similarities to obtain a similarity matrix of $15 \times 15$. In addition, the MIPL model is also used to extract the feature representations of all 15 images, and calculate the cosine similarity and similarity matrix between them through these new features. The results are shown in Figure \ref{fig:Person_Similarity_matrices}(b) (c). The experimental results show that when only the baseline is used, the large area interference due to the background, clothes, and other factors of the pedestrian image leads to a very low similarity score when the same person wears different clothes. However, when using MIPL, embedding the CIS module, BGA module, and DHP module into the baseline can effectively prompt the model to mine more identity-related cues, which in turn can extract more discriminative and more relevant features with identity information, resulting in higher similarity scores when the same person wears different clothes. In the similarity matrix, the lighter the color, the lower the correlation between elements; the darker the color, the higher the similarity. It can be seen from the pedestrian images in Figure \ref{fig:Person_Similarity_matrices}(a) that background and clothing are the areas with the greatest complexity of difference in pedestrian images, indicating that the model may pay too much attention to them, resulting in low similarity. Conversely, stable identity-related regions can reflect higher similarity, as shown by darker regions in Figure \ref{fig:Person_Similarity_matrices}(c). In addition, we also selected 15 images of different people wearing similar clothes to reflect the model's ability to identify pedestrians wearing similar clothes, as shown in Figure \ref{fig:Person_Similarity_matrices}(d). When only the baseline was used, the similarity feature diagram showed greater similarity between different pedestrians wearing similar clothing, as shown in Figure \ref{fig:Person_Similarity_matrices}(e), indicating that the extracted features may be more influenced by similar clothing information. On the contrary, when MIPL is adopted, the color blocks of the similarity feature map become lighter, as shown in Figure \ref{fig:Person_Similarity_matrices}(f), indicating that the model can effectively resist the interference of clothing appearance to a certain extent and distinguish different pedestrians wearing similar clothes. The experiment further proves the effectiveness and robustness of MIPL in focusing on the identity stable region to obtain improved similarity scores, which can better cope with the cloth-changing person ReID task.


3) To further demonstrate the effectiveness of the MIPL method, the visual search results for MIPL and baseline are shown in Figure \ref{fig:Top_10_ranking_results}, where each row is a search example, including a query image and the top 10 images that are most similar to it. It can be seen from the figure that the features extracted by the baseline method are inevitably affected by more interference factors, especially the interference of clothing texture and background. For example, in the query sample in rows (a) and (b) of Figure \ref{fig:Top_10_ranking_results}, the matching list returned by the baseline has a highly similar clothing color texture to the query sample. In contrast, the features extracted by MIPL can better resist the interference of clothing texture information, pay attention to more information with stronger identity correlation, and exceed the appearance features obtained by the baseline method. Therefore, the proposed MIPL method can potentially enhance identity invariance in specific scenarios. Moreover, for outdoor situations (Figure \ref{fig:Top_10_ranking_results}(c)(d)), higher complexity clothing textures, background information, and pose changes pose greater challenges to the model. The baseline model relies more on clothing textures and fixed pose information, and the returned results are often unsatisfactory. On the contrary, MIPL can still correctly recognize pedestrians and return more correct matching results in the top ten. These experimental results show that it is very challenging to perform the cloth-changing person ReID task when the provided images largely lack visual semantics. Moreover, the visualization results also prove that the proposed MIPL can effectively mine pedestrian identity information, and prompt learning is very helpful in overcoming the challenges of the cloth-changing person ReID task.

\section{Conclusion}

In this work, we propose a novel method named MIPL to apply to the cloth-changing person ReID task, which overcomes the complex relationship between intra-class variation and inter-class variation through the joint prompt guidance of multiple information, and explores the identity robust features from multiple perspectives. A CIS module is designed to effectively decouple the clothing information from the original image features and counteract the effect of clothing appearance. A BGA module is proposed to guide the model to learn biological key features whose identities are strongly correlated. A DHP module is constructed to minimize the impact of feature bias on model performance. Most importantly, the prompt learning processes are all jointly explored in an end-to-end unified framework. Extensive experimental results on five cloth-changing person ReID datasets verify the effectiveness of our proposed MIPL method. In particular, in terms of the accuracy of mAP and rank-1 on multiple datasets, the MIPL method outperforms the existing cloth-changing person ReID methods. Moreover, this method makes full use of various information, learns more discriminative identity robust features, and can effectively deal with the influence of interference factors such as clothing. In addition, our study also proves that the vision-language learning strategy is very helpful for solving the cloth-changing person ReID task. In the future, we intend to focus on real-person ReID scenarios and design a large-scale person ReID module that can be effectively applied to different ReID tasks, e.g., holistic person ReID, partial person ReID, occluded person ReID, and cloth-changing person ReID.


\ifCLASSOPTIONcaptionsoff
  \newpage
\fi

\normalem

\bibliographystyle{plain}
 \bibliography{mybib}

 

%




\end{document}